\documentclass{article}

\usepackage[preprint, nonatbib]{neurips_2021}

\usepackage[utf8]{inputenc} 
\usepackage[T1]{fontenc}    
\usepackage{hyperref}       
\usepackage{url}            
\usepackage{booktabs}       
\usepackage{amsfonts}       
\usepackage{amsmath}
\usepackage{nicefrac}       
\usepackage{microtype}      
\usepackage{xcolor}         
\usepackage{graphicx}
\usepackage{array}
\newcolumntype{M}[1]{>{\arraybackslash}m{#1}}
\usepackage{multirow,tabularx}

\title{Reinforcement Learning on Human Decision Models for Uniquely Collaborative AI Teammates}

\author{%
  Nicholas~Kantack\\
  Johns Hopkins University Applied Physics Laboratory\\
  Laurel, MD  20723 \\
  \texttt{nick.kantack@jhuapl.edu} \\
}

\begin{document}

\maketitle

\begin{abstract}
In 2021 the Johns Hopkins University Applied Physics Laboratory held an internal challenge to develop artificially intelligent (AI) agents that could excel at the collaborative card game Hanabi. Agents were evaluated on their ability to play with human players whom the agents had never previously encountered. This study details the development of the agent that won the challenge by achieving a human-play average score of 16.5, outperforming the current state-of-the-art for human-bot Hanabi scores. The winning agent's development consisted of observing and accurately modeling the author's decision making in Hanabi, then training with a behavioral clone of the author. Notably, the agent discovered a human-complementary play style by first mimicking human decision making, then exploring variations to the human-like strategy that led to higher simulated human-bot scores. This work examines in detail the design and implementation of this human compatible Hanabi teammate, as well as the existence and implications of human-complementary strategies and how they may be explored for more successful applications of AI in human machine teams.
\end{abstract}

\section{Introduction}

Reinforcement learning has allowed artificial intelligence (AI) systems to achieve remarkable performance in complex games from Go \cite{go} to chess and shogi \cite{chess_and_shogi} to StarCraft II \cite{starcraft}. In these (and many other) cases, the AI has learned its mastery from an arduous process of ``self-play,'' training through which the AI competes with versions of itself and explores a landscape of ever-improving strategy. While ``self-play'' can lead to extraordinary solo performances by AIs, it often fails (sometimes spectacularly \cite{l2rm}) to learn strategies that will allow it to collaborate with human teammates. The reason is simple: A self-play AI is often bad at collaborating with humans because it was trained to do something else entirely (namely, collaborate with its identical twin). An ostensible solution is to replace millions of games of ``self-play'' with millions of games of ``human-play'' (AI playing with human teammates), but an AI rarely has this kind of access to human players.

The deficiency of ``self-play'' for human machine teaming is starkly apparent in the collaborative card game ``Hanabi.'' Likened to ``team solitaire'' \cite{hanabi_challenge}, the game requires players to hold their cards face out, and the entire game revolves around giving hints and taking actions that help your teammates infer sufficient information to make successful plays. In 2019, DeepMind highlighted Hanabi as an emerging frontier for AI research \cite{hanabi_challenge}, citing the critical importance of \textit{theory of mind}, or the ability of players to monitor, model and direct the limited knowledge of others. Recently, a great deal of research has explored agents designed to excel at different types of teaming within Hanabi \cite{big_table}. A large and early portion of Hanabi agent development research has focused on producing ``self-play'' agents. In self-play, agents play with identical copies of themselves. Notably, many of these agents achieve \textit{extremely good} performance, sometimes with median scores that are perfect scores \cite{hu2021learned}. Some notable agents with high self-play scores are the Simplified Action Decoder (SAD) \cite{sad}, Rainbow \cite{rainbow}, and Fireflower \cite{fireflower}.

As remarkable as these self-play agents are, their performance outside of self-play is notably poor. Dissimilar self-play agents do not team well with each other \cite{l2rm}, revealing that there are many, non-compatible styles of excellent self-play. Furthermore, self-play agents do not play well with humans \cite{l2rm}, and are rarely made to do so (since this is somewhat of a misuse of a self-play agent). These self-play agents dominate the camp of learning-based Hanabi agents \cite{big_table}. Notably, the ``other-play'' based Hanabi agent has the best documented performance with humans to date with an average human play score of 15.8 when playing with members of a board game club \cite{other_play}. The ``other-play'' agent builds on the successes of a self-play agent (namely, SAD) but learns a strategy that is more robust to asymmetric play from its teammate.

A separate camp of Hanabi agents are rule-based agents which are constructed in a particular way to conform to human play styles and conventions \cite{big_table}. The highest human-play average score to date for rule-based agents is 15, held by intentional agents \cite{intentional_hanabi}. These agents leverage an explicit mental model of the human player along with both Hanabi-specific and generic communication conventions humans are observed to use. While rule-based agents have fallen slightly behind the latest advances in learning-based agents, rule-based agents are still regarded as better teammates by the humans with whom they play \cite{big_table}. Furthermore, rule-based agents tend to advance more sophisticated mental models of human players (i.e. a better \textit{theory of mind}), models which are highly anticipated to play significant roles in successful cases of human-machine teaming.

Designing the best Hanabi agent is challenging, since it is unclear where the upper limit of human-play scores lies, and this upper limit can vary for different groups of humans. However, pairs of humans on the online Hanabi platform \textit{Hanab.live} achieve an average score of 17 \cite{l2rm}, indicating that there is room for improvement for agents designed to play Hanabi with human teammates. To explore this opportunity, the Johns Hopkins University Applied Physics Laboratory conducted its first Hanabi challenge.

\section{The Hanabi Challenge}

In 2021, the Johns Hopkins University Applied Physics Laboratory (JHU/APL) conducted its Learning to Read Minds Challenge which tasked staff with developing Hanabi agents that would achieve high scores when paired with a single human teammate \cite{wolmetz}. SAD (``off-belief'' \cite{d_wu_off_belief}), Rainbow, and Fireflower were included in the challenge, along with four agents developed at JHU/APL. During the competition, agents would play with 21 different human players with whom the agents did not have the opportunity to train. To facilitate new research in multi-agent teaming in Hanabi, DeepMind released the \textit{Hanabi Learning Environment} (HLE) as an open source research platform \cite{hanabi_learning_environment}. This environment manages the evolution of the game state, enforces rules of the game, and creates an easy application programming interface (API) for agents and user interfaces to interact with and observe the game state. The JHU/APL challenge used the HLE for all of its competition games, and utilized a web interface for human players to play with agents within the HLE.

To determine the winner of the challenge, JHU/APL arranged for a pool of 21 human players to play two games with each of the seven agents in competition. Over the course of the competition, agents could not approach later games with knowledge from previous games (e.g. to profile human players during the competition). Rather, each agent began each competition game with no knowledge of previous competition games (sometimes considered ``zero-shot'' coordination \cite{other_play}). Furthermore, agent names were anynomized so that human players did not have any direct knowledge about the identity or style of the agent with which they played (although since the agents carried anonymized names, such as ``Agent A,'' human players could potentially play differently with an agent on the second game based on the prior game).

The remainder of this study will focus on the development of ``Cyclone,'' the winning agent. Cyclone achieved an average human-play score of 16.5, beating out the previous champion of human-play Hanabi (SAD ``off-belief'' \cite{d_wu_off_belief}) by a 4 point margin (Figure \ref{human_play}).



\section{Cyclone Agent Design}

The Cyclone agent was developed under a simple philosophy for good Hanabi play: Since humans play well with other humans, if the agent plays \textit{like} a human, it will probably play \textit{well} with humans. Therefore, the first phase of Cyclone's development focused on developing the agent itself as a model of human decision making, and the second phase focused on exploring changes to the model that allowed the agent to achieve better scores with a human teammate.

\subsection{From Factors to Decisions}

The Cyclone agent is something of a mix between a rule-based and learning-based agent. It is rule-based in the sense that is it explicitly given a low-dimensional latent space of game elements that humans pay attention to, but learning-based in the sense that it observed human play and discovered the weights humans attributed to these game elements (and weights that produce good teaming with humans). When deciding a move, the agent examines all possible actions and chooses an action that immediately (in 1 ply), maximally improves the game state according to 12 fixed \textit{factors} with learned weights. A \textit{factor} is defined as an event (e.g. playing a card, discarding a card) or quantity (e.g. number of information tokens held, change in probability of playing a playable card) that tends to impact the final score. The Cyclone agent attends to 12 factors, include factors appraising the actions of the other player, adherence to common human conventions, and managing game resources (i.e. strikes and information tokens). A complete list of factors is given in Table \ref{factor_table}, with their respective weights for different play styles the agent can employ. All version of the agent attend to these factors (or subsets thereof), but a tremendous variety of play styles can be exhibited merely by changing the weights the agent attributes to these factors.

\begin{figure}
\centering
\includegraphics[scale=0.6]{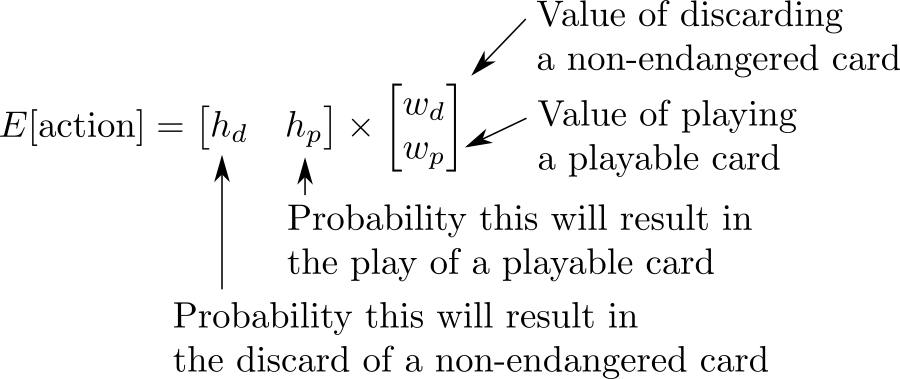}
\caption{For a simplified agent design attending to only two factors (the value of discarding and the value of playing), the calculation is shown for the expected value of an action. The expected value of the action is a sum of the values of all factors weighted by the probability that the action will cause the event the factor describes.}
\label{EV_calc}
\end{figure}

For a simple example, omit all but two of the listed factors: Value of discarding a non-endangered\footnote{An ``endangered'' card is one which has not been played and of which there is only one remaining copy in play. If one red 3 has been discarded, the remaining red 3 is endangered. Discarding an endangered card is typically bad news for the team, since it can make other cards forever unplayable (in this case, red 4s and the red 5).} card, and the value of playing a playable card. We can use these factors to build an action vector $\vec{h}$ that describes any legal game action. Specifically, the elements of the action vector specify the probability of each factor's event taking place (or the quantity by which the factor has changed, if a resource). This action vector is specific to a particular action in a particular game state. To find the expected value of an action, we take the inner product of the action's $\vec{h}$ vector with a static weights vector $\vec{w}$ that represents the agent's strategy (Figure \ref{EV_calc}).

\begin{figure}
\centering
\includegraphics[scale=0.7]{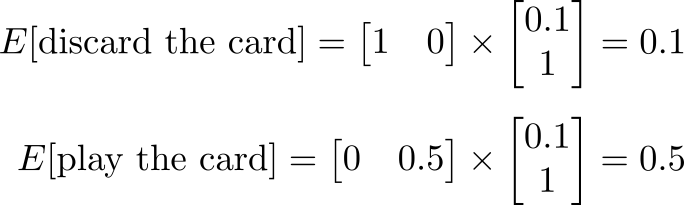}
\caption{In this example, the agent is certain (probability 1) that the card in question is not endangered, but only 50\% sure that the same card is currently playable (refer to Figure \ref{EV_calc} for the meaning of each element). These probabilities appear in the $\vec{h}$ vector and are specific to each action (since different actions have different effects), but note that the weight vector $\vec{w}$ is unchanged. Importantly, this weight vector $\vec{w}$ is static for any given version of the Cyclone agent and represents the agent's play style. In this example, the agent would choose to attempt to play the card despite uncertainty in its playability, since the expected value of playing in this circumstance is higher than the expected value of discarding (but a different weights vector $\vec{w}$ could yield a different result).}
\label{ev_examples}
\end{figure}

Suppose that the agent is certain that a particular card is non-endangered (and thus it can be safely discarded) but is only 50\% certain that the card is currently playable (i.e. there is a 0.5 probability that the card is currently playable). The associated $\vec{h}$ vectors for discarding and playing the card are shown in Figure \ref{ev_examples}, along with the calculation of each action's expected value.

If, for a set of $m$ possible actions, Cyclone calculates the corresponding factor vectors ${\vec{h}_{1}, \vec{h}_{2},..., \vec{h}_{m}}$, then the vector \begin{align}
\vec{y} = \left[\begin{matrix}
\vec{h}_{1}^{T}\\
...\\
\vec{h}_{m}^{T}
\end{matrix}\right]\vec{w}=H^{T}\vec{w}
\label{concise}
\end{align}
contains elements which represent the expected value of each action. From here, it is easy to show that the Cyclone agent's decision model amounts to a linear neural network $\left(\vec{y}=\left(I^{m \times m} \otimes \vec{w}^{T}\right)\left(\textnormal{vec}(H)\right)\right)$ with sparse weights built up from $\vec{w}$.

\subsection{Enumerating the Twelve Factors}

The list of factors to which the Cyclone agent attends was built over time in consultation with intermediate Hanabi players to progressively increase the humanness of the agent's play style. The factors fall into three categories (Play, Discard, and Conventions), and will now be defined. These factors (along with their values for different play styles) were introduced in a prior publication \cite{iai}, and are summarized in the appendix (Table \ref{factor_table}).

\textbf{Play.} There are five factors that relate to the value (or penalty, if negative) associated with attempts (successful and unsuccessful) to play a card. These factors are: playing a playable card, misplaying\footnote{``Misplaying'' is defined as attempting to play a card that is current not playable. This incurs a strike, and ends the game if it incurs the third strike.} (less than 2 strikes), misplaying (2 strikes), the other player playing a card, and the other player misplaying.

\textbf{Discard.} There are two factors that relate to the value of discarding a card. Unlike the \textit{Play} factors, there are no explicit penalties for discarding (even though some discards would be very unfortunate), except one mentioned in the \textit{Conventions} factors. These factors are: discarding a non-endangered card$^{1}$, and discarding an unneeded\footnote{An ``unneeded'' card is a card that cannot be played (e.g. the red 4 is unneeded if the other red 4 has been played, but also if all of the red 3s have been discarded).} card.

\textbf{Conventions.} There are five factors associated with a common convention among humans concerning singling out cards. A clue is said to ``single out'' a card if the receiving player learns information about only one of the cards mentioned in the clue. The clue ``this is your only red card'' singles out the card specified, but the Cyclone agent also considers the clue ``these are your two red cards'' to single out a card if one of the two cards was already known to be red (in which case, the card that was not known to be red was singled out). Humans most often single out playable cards, an indeed experiments with the agent revealed that allowing this convention between players leads to a significant improvement in scores. Importantly, these factors only apply to one's own actions. Therefore, the agent does not make an effort to predict the likelihood of its teammate following this convention. These factors are: playing a singled out card, discarding a singled out card, giving a clue that singles out a playable card, giving a clue that singles out a non-playable card, and giving a clue when information tokens are held. This final factor (``giving a clue when information tokens are held'') is the only factor that is not numerically represented as the probability of an event taking place, but rather, is equal to the number of information tokens held if the player is giving a clue, otherwise its value is zero. The purpose of the information token factor is to allow an agent to prioritize spending down information tokens near the end of the game (since unspent information tokens at the end of the game are wasted).

\subsection{Mental Model of a Human Player}

Some of the 12 factors involve actions taken by the other player, so a model of the other player's decision making is needed in order to estimate the probability of the other player taking certain actions. The Cyclone agent models its human teammate as a self-copy with different \textit{knowledge} and \textit{weights}. To estimate the probability that its teammate will play a card, the agent adopts the following probabilistic model. The probability that one's teammate will play their $i$th card on the next turn is \textit{assumed to be equal to} the probability (from one's teammate's perspective) that their $i$th card is playable. Similarly, the probability that one's teammate will discard their $i$th card on the next turn is \textit{assumed to be equal to} the probability (from one's teammate's perspective) that their $i$th card is non-endangered. In particular, one's teammate is assumed to take into account all information available (e.g. past clues, discarded cards, etc.) in evaluating this probability. If $P_{\textnormal{\small play }i}$ is the probability that one's teammate will play their $i$th card, then the model holds that

\begin{align}
P_{\textnormal{\small play } i} = \frac{\textnormal{\# of unseen, playable cards matching clues for card }i}{\textnormal{\# of unseen cards matching clues for card }i}
\end{align}

It is worth noting that this probabilistic model does not satisfy the probability simplex. For example, if the Cyclone agent's first action in the game is to inform the other player that they hold a 1, the agent's model for its teammates holds that both the probability the teammate will play the card and the probability the teammate will discard the card are equal to 1. However, playing and discarding are mutually exclusive actions, so the model cannot be taken as a true probability distribution. Functionally, however, this does not present a problem, since any necessary normalization will naturally be absorbed in the weights during training.

\subsection{``Giving Up'' on Cards}

Many agents (and humans) will play Hanabi in a way that does not obviously reduce the maximum possible number of points. For instance, human players will often try to avoid discarding a 5 because doing so necessarily reduces the maximum possible attainable score (since before the discard, that 5 might later have been played, but after being discarded it can never be played). However, the Cyclone agent discovered that when the goal is to maximize the expected score (not the maximum score possible), it can be beneficial to ``give up'' on playing certain cards if it appears unlikely that the opportunity to play those cards will arise. This is also intuitive; if there are only 8 turns remaining and no green cards have been played, it is very unlikely that \textit{all} green cards will be played before the end of the game, and so it might be beneficial to discard the green 5 in the hopes of maximizing the number of cards that are played in the final 8 moves.

\begin{figure}
\centering
\includegraphics[scale=0.65]{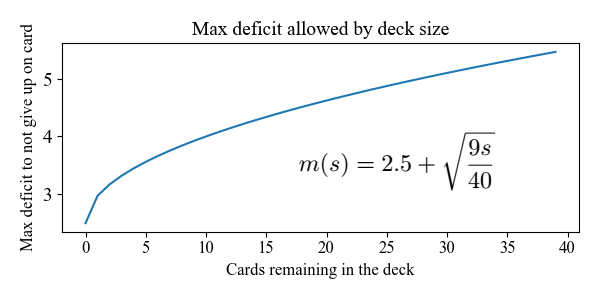}
\caption{If the deficit for a given card exceeds the value given by this curve (for the current deck size), the card is considered unneeded (discarding the card is worth more than keeping it in the hopes of playing it). In the equation, $m$ is the height of the curve, and $s$ is the deck size at the moment of calculation.}
\label{give_up}
\end{figure}

The decision to give up hope on playing a card is based on that card's \textit{deficit}. The \textit{deficit} for a card is the card's rank minus the highest rank that has been played for that color. If a green 2 has been successfully played, the deficit for a green 4 is 2 (4 (the green 4 in question) $-$ 2 (the highest green rank played) = 2 (the deficit)). If the deficit exceeds a certain limit, the Cyclone agent considers the card ``unneeded'' (i.e. the agent believes the value of discarding the card exceeds the expected value of trying to keep it around to play in the future). One can imagine that a good deficit threshold would be sensitive to the size of the deck (e.g. if the game has just started, there can be high hopes for overcoming a high deficit. If the game is nearly over, playing a high deficit card becomes more unlikely). The best maximum deficit curve the agent found during training is displayed in Figure  \ref{give_up}. This model indicates that at the beginning of the game ($s=40$) a deficit of 5.5 is tolerated, so no card will be given up until the deck size reaches 29. At this stage, 5s will freely be discarded for any color for which a 1 has not been played. This critical deficit continues to taper over the course of the game, placing increasing focus on immediately playable cards.

\section{Cyclone Development and Training}

As stated before, the Cyclone agent is both a ruled-based agent and a learning-based agent. The previous sections describe how the agent's attention to 12 fixed factors constitute the rule-based aspects of the agent's design. However, the weights given to these factors dramatically influence the ultimate play style, and these weights are learned quantities. Prior to learning, the weights were initialized to best-guess values by consulting intermediate Hanabi players. The learning process involved gradient estimation via a sequence of full-factorial design experiments. For each experiment, four weights were selected. Each selected weight was decreased, held constant, or increased (3 levels), and all combinations of levels across the four weights were tested ($3^{4}=81$ weight vectors tested per experiment). The weight vector leading to the highest scores served as the starting point for the next experiment. These experiments continued until they saturated (i.e. the highest performing weight vector was the starting point of the experiment).

\begin{figure}
\centering
\includegraphics[scale=0.7]{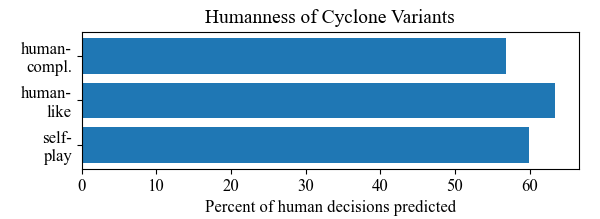}
\caption{The ``humanness'' of the Cyclone agent variants is shown. This is the fraction of games states (from a database of 376 human decisions and game states) in which the variant recommended the same action as the action taken by the human (the author). The highest humanness of any agent was 64.2\%, achieved by the human-like agent.}
\label{humanness}
\end{figure}

\begin{figure}
\centering
\includegraphics[scale=0.65]{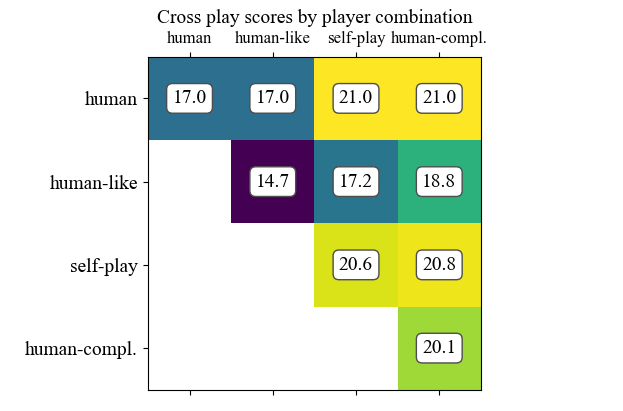}
\caption{The mean score is shown for different pairings of the Cyclone agent variants (and humans). Note that the pool of humans sampled for these data (2 humans) is very small, and these human play results are not the same as those shown in Figure \ref{human_play}. These data were collected earlier, during the development of the Cyclone agent. As expected, the human-complementary agent achieved the highest score when paired with the human-like agent out of all agents shown, since this is the specific task for which human-complementary agent was optimized. 95\% error margins are $\pm$ 0.3 point (except for scores involving human play). Notably, several instances of asymmetric play (e.g. human + human-compl., self-play + human-compl.) lead to better results than the self play of either member of the asymmetric pairing. Indeed, the three highest average scores shown are for asymmetric pairings, indicating that optimization outside of self-play can be beneficial for generic teaming.}
\label{cross_play}
\end{figure}

A small database of 376 human decisions (made by the author while playing Hanabi) served as the basis for learning weights that led to human-like play. The design experiments results were sorted on their ability to produce a weight vector which led to the highest fraction of human decisions being predicted by the agent. This led to the ``human-like'' version of the Cyclone agent which served as a behavior clone for human players. A second version of the agent was generated through this learning process using self-play scores as the cost function (leading to the ``self-play'' agent). Finally, this process was repeated for a third version of the agent that was trained to achieve high scores when paired with the human-like version of the agent (producing the ``human-complementary'' version). The humanness of each agent version is shown in Figure \ref{humanness}, and the average scores are shown in Figure \ref{cross_play} for different combinations of agents.

\section{Results}

\begin{figure}
\centering
\includegraphics[scale=0.7]{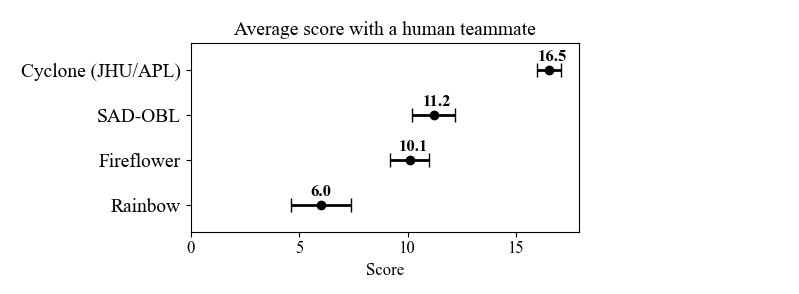}
\caption{The average scores are shown for SAD ``off-belief'' \cite{d_wu_off_belief}, Rainbow \cite{rainbow}, Fireflower \cite{fireflower}, and Cyclone. These scores were obtained by pairing an agent with a human teammate (drawn from a pool of 21). The error bars show the 95\% confidence interval of the mean. At least 49 games were included for each agent shown. More detail on results are available in \cite{l2rm}.}
\label{human_play}
\end{figure}

The human-complementary version of the Cyclone agent won JHU/APL's Learning to Read Minds Challenge with a human-play score of 16.5, higher than the highest human-play scores reported for Hanabi agents to date \cite{big_table}\cite{intentional_hanabi}. This score is shown alongside those for other benchmark agents in Figure \ref{human_play} (including the SAD ``off-belief'' agent, considered the state-of-the-art for human teaming Hanabi agents \cite{big_table}\cite{d_wu_off_belief}). The Cyclone agent's success over self-play agents (e.g. Rainbow) is expected, since the competition version of the agent was explicitly optimized for play with a human behavioral clone (the human-like Cyclone agent). The agent's rule-based design proved beneficial for the purposes of making an (64\%) accurate human decision model (the human-like agent) with which to train. Furthermore, the agent's strategy can clearly be interpreted in human-understandable terms (i.e. weights on human-understood factors). In this way, the Cyclone agent was able to learn insight into human play that can be leveraged both for human decision modeling as well as explanations (including a novel means of AI ``instruction'' \cite{iai}). Importantly, these results demonstrate that the Cyclone agent can be a strong ally for a human teammate, even perhaps a \textit{better} ally than a human, evidenced by the higher average score for human-agent pairs vs. human-human pairs (Figure \ref{cross_play}).

A significant result from the experiments combining different versions of the Cyclone agent is that the top asymmetric teams (teams not comprised of two identical copies of an agent) achieved higher average scores than the top symmetric teams. For instance, pairing the human-complementary agent with a human led to higher scores than either member's self play (17 for human + human, 20.1 for human-compl. + human-compl.). Additionally, pairing the human-complementary agent with the self-play agent led (quite interestingly) to a higher average score (20.8) than either self-play pair (20.6 for self-play + self-play, 20.1 for human-compl. + human-compl.). These results indicate that for the factor-based Cyclone architecture, \textit{self-play might hinder rather than help teaming}. These results indicate that asymmetric, complementary strategies open a significantly wider (and indeed better performing) state space of possible teams. There is good evidence in the Cyclone agent cross play data to support this notion. In particular, the human-complementary agent is \textit{more} likely to single out playable cards than the self-play agent, but simultaneously \textit{less} likely to assume a singled out card is playable (because humans are more prone to single out non-playable cards, as evidenced by the human-like agent's much smaller penalty for singling out non-playable cards).

Examining the differences in the weights between the human-like agent and the human-complementary agent sheds some light on why the human-complementary agent excels when playing with humans. The human-complementary agent's discard weights are much higher than those of the human-like agent (Table \ref{factor_table}), indicating that the human-complementary agent has learned to discard more often so that its human teammate will discard less often, likely because the Cyclone agent is a better judge of what to discard. Consistent with this hypothesis, the human-complementary agent places six times as much weight on making sure its teammate (nominally a human) knows which cards are safe to play.

While the Cyclone agent achieved high scores with human teammates, there are many aspects of the agent that have not been thoroughly optimized. First, the 12 factors were identified through the speculation of a small number of Hanabi players. Consequently, there is nothing particularly optimal about the factors. Developing techniques to discover factors organically would greatly extend the versatility (and likely optimality) of an agent employing this particular architecture. Additionally, there is potential for much more robust behavioral clone development. The development of the Cyclone agent leveraged only a single behavioral clone emulating a single observed human player (the author). Future work should explore the advantages of generating multiple behavioral clones modeled after multiple humans, perhaps using an instruction-based modeling process \cite{iai}. Exploring these opportunities for improvement on the agent development may reveal powerful new techniques and optimization processes that will allow artificially intelligent agents to learn highly effective strategies for collaborating with human teammates.

\section{Conclusion}

The JHU/APL Learning to Read Minds Challenge saw the development of a Hanabi agent that plays very well with human teammates, achieving record human-play scores. These results were made possible with a hybrid rule-based, learning-based approach to agent development that allowed the creation of highly accurate behavioral clones for human players with which the agent could train. This training process gave rise to definitive evidence for uniquely human complementary play styles and clearly demonstrated the existence of highly successful asymmetric strategies that cannot be discovered through traditional self-play optimization techniques. Optimization processes built around techniques to simulate human teaming show significantly better human teaming results than traditional self-play training. These new optimization methods will likely serve as the basis for the development of successful AI agents who must work collaboratively with humans to achieve common goals.

\small
\bibliographystyle{unsrt}
\bibliography{ref}

\newpage
\appendix

\section{Appendix}

\begin{table}[h]
  \caption{The twelve factors, along with their values for three different Cyclone play styles (human-like, human-complementary, and self-play)}
  \label{factor_table}
  \centering
  \begin{tabularx}{\textwidth}{lllll}
\toprule
Category & Factor & human-like     & human-compl. & self-play\\
\midrule
\multirow{5}{*}{\rotatebox{0}{Play}} & Playing a playable card & 1 & $\infty$ & 11\vspace{0.5em}\\
& Playing an unplayable card\\
& (fewer than 2 strikes) & -1  & -1 & -1\vspace{0.5em}\\
& Playing an unplayable card (2 strikes) & 3 & $\infty$ & 3\vspace{0.5em}\\
& Other player playing a\\
& playable card & 1.5 & 10 & 2\vspace{0.5em}\\
& Other player playing an\\
& unplayable card & 0 & 0 & 1\vspace{0.5em}\\
\hline\\
\multirow{2}{*}{\rotatebox{0}{Discard}}& Discarding a non-endangered card & 0.1 & 0.55 & 0.8\vspace{0.5em}\\
& Discarding an unneeded card & 0.25 & 1 & 0\vspace{0.5em}\\
\hline\\
\multirow{5}{*}{\rotatebox{0}{Conventions}} & Playing a singled out card & 3 & 1.5 & 5\vspace{0.5em}\\
& Giving a clue that singles\\
& out a playable card & 3 & 3 & 2\vspace{0.5em}\\
& Giving a clue that singles\\
& out a non-playable card & 0 & -5 & -4\vspace{0.5em}\\
& Discarding a singled out card & -0.5 & -2 & -3\vspace{0.5em}\\
& Added value to any clue\\
& per info token held & 0.5 & 0.1 & 0\vspace{0.5em}\\
\bottomrule
\end{tabularx}
\end{table}

\end{document}